\documentclass{article}

\pdfoutput=1

    \usepackage[preprint]{neurips_2024}

\usepackage[utf8]{inputenc} 
\usepackage[T1]{fontenc}    
\usepackage{hyperref}       
\usepackage{url}            
\usepackage{booktabs}       
\usepackage{amsfonts}       
\usepackage{nicefrac}       
\usepackage{microtype}      
\usepackage{xcolor}         

\usepackage{colortbl}

\definecolor{fst}{rgb}{1.0, 0.7, 0.7}
\definecolor{snd}{rgb}{1.0, 0.85, 0.7}
\definecolor{trd}{rgb}{0.95, 0.95, 0.65}

\definecolor{hsy}{rgb}{0.0, 0.0, 0.0}
\newcommand{\hsy}[1]{{\color{hsy} #1}}

\definecolor{fjh}{rgb}{0.0, 0.0, 0.0}
\newcommand{\fjh}[1]{{\color{fjh} #1}}

\definecolor{fjhI}{rgb}{0.0, 0.0, 0.0}

\definecolor{fjhcomment}{rgb}{0.0, 0.0, 0.0}

\definecolor{thy}{rgb}{0.0, 0.0, 0.0}
\newcommand{\thy}[1]{{\color{thy} #1}}

\definecolor{ZouColor}{rgb}{0.0,0.0,0.0} 
\definecolor{ZouColor2}{rgb}{0.0,0.0,0.0} %

\newcommand{\Z}[1]{{\color{ZouColor2} #1}}

\usepackage{microtype}

\usepackage[normalem]{ulem}
\useunder{\uline}{\ul}{}

\usepackage[pdftex]{graphicx}
\usepackage{amsmath} 

\newcommand{\Shi}{\cite{shi2020motionet}}
\newcommand{\Ma}{\cite{ma2021context}}
\newcommand{\Gong}{\cite{gong2021poseaug}}
\newcommand{\Li}{\cite{li2021human}}
\newcommand{\Zhan}{\cite{Zhan_2022_CVPR}}
\newcommand{\Geng}{\cite{Geng23PCT}}
\newcommand{\Ci}{\cite{ci2023gfpose}}

\usepackage{bm}

\title{Learning Human Motion from Monocular Videos\\
via Cross-Modal Manifold Alignment}

\author{
    Shuaiying Hou \\ State Key Lab of CAD\&CG \\ Zhejiang University \\
    \And
    Hongyu Tao \\ State Key Lab of CAD\&CG \\ Zhejiang University \\
    \And
    Junheng Fang \\ Bournemouth University \\
    \And
    Changqing Zou \\ State Key Lab of CAD\&CG \\ Zhejiang University \\
    \And
    Weiwei Xu\thanks{Corresponding Author} \\ State Key Lab of CAD\&CG \\ Zhejiang University
}

\begin{document}
\maketitle
\begin{abstract}
Learning 3D human motion from 2D inputs is a fundamental task in the realms of computer vision and computer graphics. Many previous methods grapple with this inherently ambiguous task by introducing motion priors into the learning process. However, these approaches face difficulties in defining the complete configurations of such priors or training a robust model. In this paper, we present the Video-to-Motion Generator (VTM), which leverages motion priors through cross-modal latent feature space alignment between 3D human motion and 2D inputs, namely videos and 2D keypoints. To reduce the complexity of modeling motion priors, we model the motion data separately for the upper and lower body parts. Additionally, we align the motion data with a scale-invariant virtual skeleton to mitigate the interference of human skeleton variations to the motion priors. Evaluated on AIST++, the VTM showcases state-of-the-art performance in reconstructing 3D human motion from monocular videos. Notably, our VTM exhibits the capabilities for generalization to unseen view angles and in-the-wild videos.
\end{abstract}    
\section{Introduction}
\label{sec:intro}
%
\hsy{Modeling 3D human motion from 2D inputs, e.g. 2D keypoint sequences or monocular videos, is a fundamental cornerstone for various computer vision and computer graphics tasks, such as human action recognition, behavior analysis, video games, and virtual/augmented reality, rtc. However, this endeavor presents considerable challenges due to the inherent ambiguities in inferring 3D from 2D. Imposing motion priors has proven an effective strategy to mitigate ambiguities and increase the 3D pose plausibility.}

\hsy{Previous works on modeling motion priors can be broadly categorized into two groups: explicit and implicit \Z{prior based methods}. Explicit methods primarily focus on restricting joint angles based on biomechanics~\cite{HATZE1997128,1043949} or directly optimizing observed data~\cite{7298751}. Unfortunately, the full configuration of pose-dependent joint angle limitations for the whole body remains unknown. Implicit methods include those modeling the plausible motion distribution with Gaussian mixture model~\cite{bogo2016keep, hou2023causal}, variational autoencoders (VAEs)~\cite{SMPL-X:2019,10.1145/3386569.3392422}, generative adversarial networks (GANs)~\cite{10.1145/3450626.3459670,9879900} or denoising score matching (DSM)~\cite{ci2023gfpose}. These models either suffer from issues like posterior/mode collapse or \Z{difficulty in training}.}

\begin{figure}[t]
  \centering
  \includegraphics[width=0.98\linewidth]{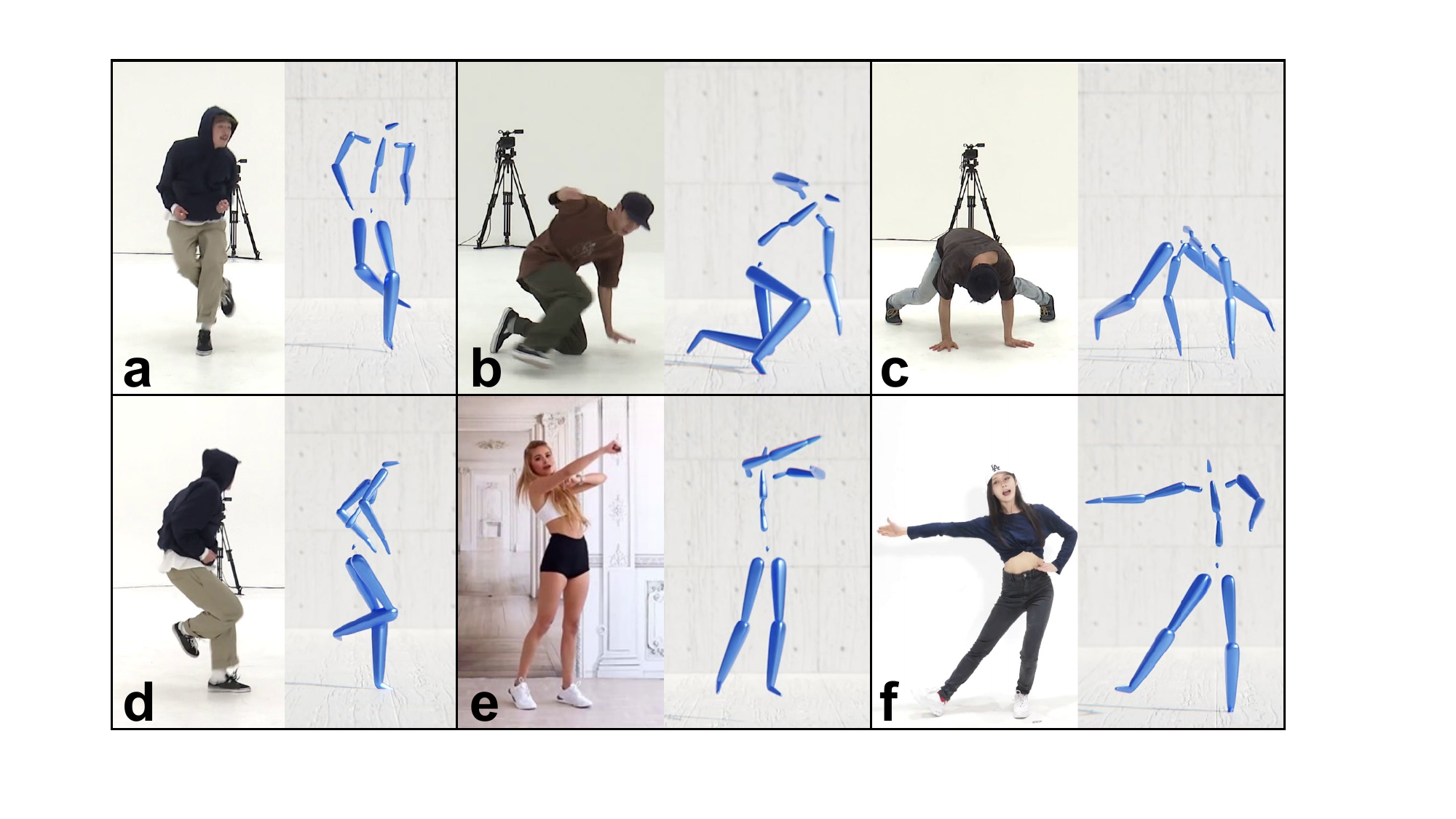}
  \caption{Our VTM can reconstruct 3D human motion from a wide range of} monocular videos. These poses can be reconstructed from: a), b) and c), three frames from the validation dataset, d) the frame containing the same pose as that in a) but with an unseen view-angle during training, e) and f), two frames from the internet videos.
  \label{fig:teaser}
\end{figure}

\hsy{In light of the success of the contrastive language-image pre-training (CLIP) model~\cite{radford2021learning}, we take a new perspective in harnessing 3D motion priors to aid the reconstruction of human motion from monocular videos by aligning the latent feature space of 3D skeleton motion data and video data. As several datasets, such as ~\cite{li2021learn,ionescu2013human3}, contain the 2D videos of human performance and their corresponding 3D motion, it's possible to find a compact and well-defined latent manifold shared by data from these two domains, similar to the CLIP latent manifold. This latent space should contain ample information to reconstruct human poses faithfully. Therefore, there are two technical challenges when performing cross-modal alignment in 3D human pose reconstruction: 1) identifying such a latent manifold, which encapsulates the kinematic constraints embodied in 3D human motion and serves as our motion priors, and 2) finding an effective way to align video feature space to the motion priors to achieve high-fidelity 3D motion reconstruction.}


\hsy{To address the aforementioned two challenges, we introduce a novel neural network called VTM. Firstly, the VTM employs a two-part motion auto-encoder (TPMAE), which segregates the learning of the motion latent manifold into the upper and lower body parts. This division effectively reduces the complexity of modeling the entire human pose manifold. The TPMAE is trained on 3D human motion data aligned with a scale-invariant virtual skeleton, which is beneficial for eliminating the impact of the skeleton scale on the manifold. Subsequently, the VTM incorporates a two-part visual encoder (TPVE) to translate visual features, composed of video frames and 2D keypoints, into two streams of visual latent manifolds for the upper and lower body parts. A manifold alignment loss is employed to pull the cross-modal manifolds for the upper and lower body parts closer. Finally, the TPVE is jointly trained with the pre-trained TPMAE to reconstruct 3D human motion with a complete representation: the rotations for all joints, the translations for the root joint, and a kinematic skeleton containing the scale information. This meticulously designed process ensures harmonization between the motion and visual representations within the VTM framework.
} 

\hsy{We comprehensively evaluate our VTM framework on the AIST++ dataset~\cite{li2021learn}. The results demonstrate that our VTM surpasses or performs comparably to state-of-the-art (SOTA) methods in terms of mean per joint position error (MPJPE). Notably, VTM exhibits the capability to swiftly generate precise human motion in sync with video sequences (\textasciitilde 70fps). This real-time performance positions it as a potential candidate for various real-time applications. In summary, the contributions of this paper are as follows:
\begin{itemize}
    \item We introduce a novel neural network, VTM, for learning human motion from video sequences. Our experimental results verify VTM's ability to faithfully replicate motion recorded in the provided videos.
    \item We present a new approach for harnessing scale-independent 3D motion priors by aligning motion data and video data on the two-body-part latent feature manifolds. This approach has proven effective in generating high-quality motion in our experiments.
    \item VTM produces a complete motion representation consisting of a kinematic skeleton, reliable global root translations, and naturally accurate joint rotations, thereby enabling more versatile applications in character motion and motion analysis.
\end{itemize}
}

\section{Related Works}
\label{sec:relatedworks}
\fjh{Human Pose Estimation (HPE) is a fundamental challenge in computer vision that plays a crucial role in various advanced applications. In simple terms, it involves the precise determination of the position and orientation of pivotal anatomical keypoints, such as the head, left hand, right foot, etc. when given a video, a signal, or an image. With the rapid growth of deep learning techniques, there has been a wealth of methods developed for 2D HPE in~\cite{andriluka20142d,chen20222d,munea2020progress,dang2019deep,zheng2023deep} and 3D HPE in~\cite{sarafianos20163d,wang2021deep,chen2020monocular}. Furthermore, an array of methodologies pertaining to human body representation including SCAPE~\cite{anguelov2005scape}, SMPL~\cite{loper2015smpl}, skeleton-based model~\cite{cao2018openpose}, surface-based model~\cite{guler2018densepose}, and Pose as Compositional Tokens (PCT)~\cite{geng2023human}, as well as accompanying datasets, including HumanEva-I~\cite{sigal2010humaneva}, Human3.6M~\cite{ionescu2013human3}, MARCOnI~\cite{elhayek2016marconi}, Microsoft COCO~\cite{lin2014microsoft}, etc.. have been proposed aiming to advance the field. In this section, we delve into three main paradigms within the forefront of 3D human pose estimation techniques: direct 3D HPE in Sec.~\ref{sec:direct3d}, lifting from 2D to 3D pose in Sec.~\ref{sec:lifting3d}, and optimization \& regression-based methods in Sec.~\ref{sec:or3d}.}

\begin{figure*}[t]
  \centering
  \includegraphics[width=0.98\linewidth]{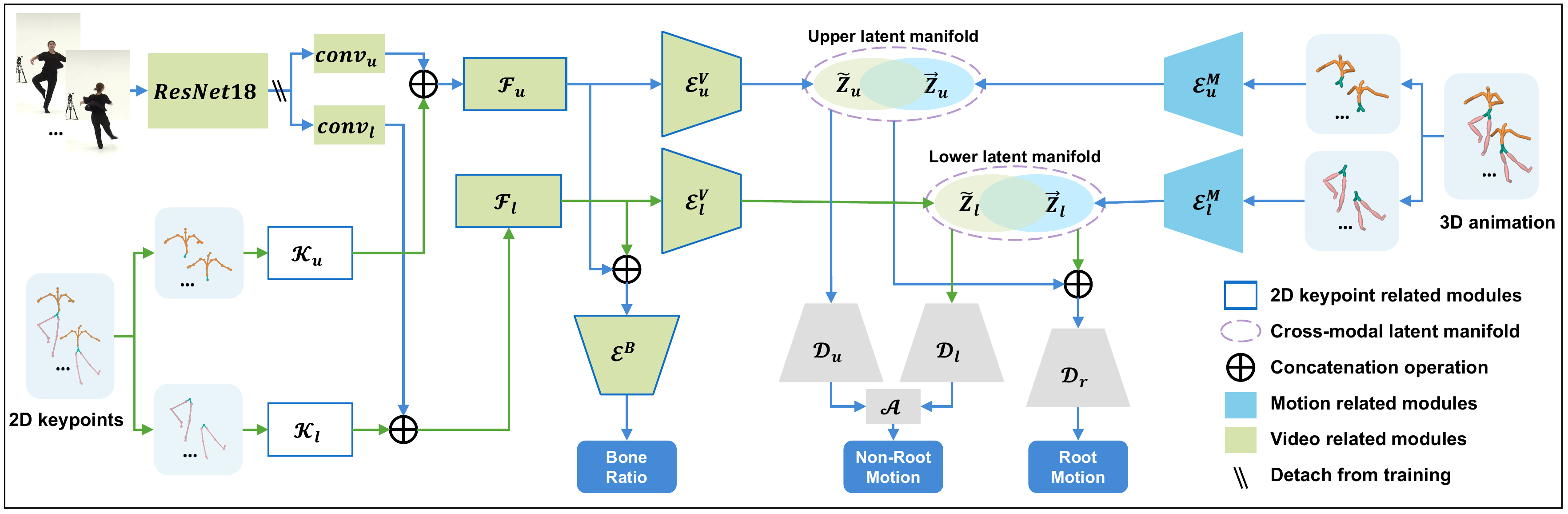}
  \caption{System overview of our VTM. The TPMAE (including motion encoders $\mathcal{E}_u^M$ \& $\mathcal{E}_l^M$, motion decoders $\mathcal{D}_u$ \& $\mathcal{D}_l$, and root decoder $\mathcal{D}_r$) are first trained on the motion data to learn two latent manifolds as the motion priors. Then, the TPVE (including 2D keypoints feature extractors $\mathcal{K}_u$ \& $\mathcal{K}_l$, visual fusion blocks $\mathcal{F}_u$ \& $\mathcal{F}_l$, visual encoders $\mathcal{E}_u^V$ \& $\mathcal{E}_l^V$ and bone ratio predictor $\mathcal{E}^B$) are jointly trained with the pre-trained TPMAE to align the visual features with the motion priors for reconstructing 3D human motion. The superscripts $M$ and $V$ represent the ``Motion'' and ``Video''; the subscripts $u$, $l$ and $r$ mean the ``upper body part'', ``lower body part'' and ``root''.}
  \label{fig:overview}
\end{figure*}

\subsection{Direct 3D HPE Methods}
\label{sec:direct3d}
\fjh{Similar to the approach employed in 2D HPE, the most straightforward method for 3D human pose estimation involves the utilization of an end-to-end neural network to make predictions regarding the 3D joint coordinates. Within this context, detection-based methods~\cite{pavlakos2017coarse,luvizon20182d} predict a likelihood heatmap to determine the location of each joint, whilst regression-based methods treat the estimation of the joint locations relative to the root joint as a regression problem~\cite{li20153d,zhou2016deep,sun2017compositional,luvizon2019human,nibali2018numerical}. In a notable development, Sun et al.~\cite{sun2018integral} innovatively integrated the regression approach in the joint position estimated from the heatmap, which effectively reduces the computational and storage cost associated with the end-to-end training process.}

\subsection{Lifting from 2D to 3D Pose}
\label{sec:lifting3d}
\fjh{Given the unequivocal one-to-one correspondence between 2D and 3D joints, 3D HPE could benefit from the 2D HPE results as a means to enhance its in-the-wild generalization performance. For instance, utilizing a simple neural network to learn the lifting from 2D to 3D pose, which was popularized by the work in~\cite{martinez2017simple}. These works~\cite{park20163d,tekin2017learning,habibie2019wild,zhou2019hemlets} have made significant progress in addressing the inherent challenge by tackling the fusion of 2D heatmaps with 3D image cues. Other methods, including long short-term memory (LSTM)~\cite{nie2019single,wang2019not}, Euclidean distance matrix~\cite{moreno20173d}, and graph neural networks ~\cite{zhao2019semantic,ci2019optimizing}, have leveraged the correspondence between joints for designing advanced algorithms. During training, it is often to employ 2D space reprojections of the 3D pose as a form of supervision to ensure consistency~\cite{tome2017lifting,habibie2019wild,chen2019unsupervised}. 
The Transformer model~\cite{vaswani2017attention} additionally furnishes an alternative avenue for optimizing the transformation process from 2D representations to their corresponding 3D counterparts~\cite{einfalt2023uplift,li2022exploiting,li2022mhformer,shan2022p,zhang2022mixste,tang20233d}.
{Zheng et al.~\cite{zheng20213d} have proposed PoseFormer, which utilizes cascaded transformer layers to achieve remarkable performance, and PoseFormerV2~\cite{zhao2023poseformerv2}, where enhancements in both efficiency and robustness have been achieved.} 
Furthermore, with the introduction of GANs(Generative Adversarial Networks), more realistic 3D human poses have been generated with the help of a discriminator~\cite{fish2017adversarial,yang20183d,wandt2019repnet}.}

\subsection{Optimization and Regression-based Methods}
\label{sec:or3d}
\fjh{
In recent times, the skinned multi-person linear (SMPL) model~\cite{loper2015smpl} has become a popular choice for the representation of human bodies when fitting to images employing annotated keypoints and silhouettes. Many optimization techniques tailored for the SMPL model were introduced for 3D HPE~\cite{bogo2016keep,lassner2017unite,han2023licamgait,liang2023hybridcap} due to its inherent capacity to incorporate prior knowledge concerning human body shape, rendering it amenable to effective fitting even with limited training data. However, optimization-based approaches relying solely on these fitting techniques may lead to unnatural body shapes and poses. Some other methods regress the SMPL parameters by various networks~\cite{kanazawa2018end,pavlakos2018learning,omran2018neural,zhang2021body,sun2021monocular,sun2022putting} due to their effectiveness, whilst regression-based approaches encounter challenges in maintaining precise alignment between images and the corresponding mesh representations.
Moreover, the Inverse Kinematics Optimization Layer (IKOL)~\cite{zhang2023ikol} utilizes the SPIN (SMPL oPtimization IN the loop)~\cite{kolotouros2019learning} framework to amalgamate optimization-based and regression-based methodologies, whose synergy leads to a marked improvement in performance.}

In this paper, we focus on reconstructing the skeleton, the joint rotations, and the global root translations from the video features and 2D poses by aligning them with the pre-learned motion priors.

\section{Video-to-Motion Generator}
\subsection{Data Preparation}
\label{sec:data_preparation}
\hsy{
\textbf{Dataset.} A cross-modal dataset containing both 3D joint rotation data and 2D videos, like the AIST++~\cite{li2021learn} and Human3.6M~\cite{ionescu2013human3} dataset, is indispensable for our objective of generating a complete motion representation for diverse characters from monocular videos. The AIST++ dataset contains far more performers than the Human3.6M dataset (27 vs. 7), which means there are more skeleton variations in AIST++. This makes it more challenging to model the motion in AIST++ but more suitable for our task since we want to reconstruct the skeletons from the 2D inputs. Therefore, we opt for the AIST++ dataset to train and evaluate our VTM. Our experiments only utilize the 2D keypoints and videos captured by camera 1 as our single-view 2D input; more details about the camera settings can be found in~\cite{aist-dance-db}.
}

\hsy{\textbf{Data Preprocess.} Firstly, we inspect the dataset, excluding sequences that exhibit incorrect poses, to ensure the learning of a well-defined latent manifold. Subsequently, leveraging the 3D joint positions, we construct $J$-joint skeletons, adjusting bone lengths to match the characters in the videos. Finally, employing the joint rotations and the created skeletons, we generate motion files in BVH format, resulting in the BVHAIST++ dataset. This dataset will be publicly available alongside our source code.

Following this, we construct a virtual skeleton, denoted as $\mathbf{\bar{s}}$, by averaging the skeletons present in BVHAIST++. All motion data undergo alignment with $\mathbf{\bar{s}}$, yielding a skeleton-scale-independent motion dataset. This dataset facilitates the learning of motion priors disentangled from the specifics of the skeleton scale. Additionally, we define the bone ratios $\mathbf{b}$ as the ratios between the original skeletons and the average skeleton. During inference, the ratios predicted by VTM are used to recover the original skeletons from the averaged skeleton.
} 


\hsy{
\textbf{3D Motion Representation.} We convert the BVH files of $\mathbf{\bar{s}}$ into camera space using the camera parameters and represent the $t_{th}$ frame of the transformed motion sequence as $\mathbf{x}_t=\{ \mathbf{r}_t^q, \mathbf{r}_t^p, \mathbf{r}_t^v, \mathbf{x}_t^q, \mathbf{x}_t^p, \mathbf{x}_t^v \}$. Here, $\mathbf{r}_t^q \in \mathbb{R}^{6}$ and $\mathbf{x}_t^q \in \mathbb{R}^{6(J-1)}$ denote the 6D rotation representations~\cite{Zhou_2019_CVPR} of the root and non-root joints. $\mathbf{r}_t^p \in \mathbb{R}^{3}$ and $\mathbf{x}_t^p \in \mathbb{R}^{3(J-1)}$ represent the global 3D joint positions, while $\mathbf{r}_t^v \in \mathbb{R}^{3}$ and $\mathbf{x}_t^v \in \mathbb{R}^{3(J-1)}$ are the linear velocities.
}

\hsy{
\textbf{2D Input Representation.} To maintain scale consistency between 2D keypoints and 3D motion, we project the 3D joint positions obtained through the forward kinematic (FK) process into 2D utilizing the camera parameters. These 2D keypoints are referred to as virtual 2D keypoints. Then, we represent the $t_{th}$ frame of the 2D keypoints input as $\mathbf{k}_t = \{ \mathbf{k}_t^p, \mathbf{k}_t^v \}$, where $\mathbf{k}_t^p \in \mathbb{R}^{2J}$ represents the 2D virtual keypoints, and $\mathbf{k}_t^v \in \mathbb{R}^{2J}$ corresponds to their linear velocities.
}

\subsection{Learn Motion Priors}
\label{sec:learn_mp}
\hsy{
To learn compact and well-defined priors from the motion data, we design a two-part motion auto-encoder (TPMAE) comprised of the motion encoders, motion decoders, and root decoder depicted in Fig.~\ref{fig:overview}. TPMAE integrates the two-part approach, inspired by the works of~\cite{hou2023tptn,10.1145/3610548.3618181}, into a convolutional architecture closely resembling the one employed in~\cite{zhang2023generating} (the details can be found in the supplementary material).}

\hsy{\textbf{Encoders.} Given an input sequence comprising $T$ frames, denoted as $\mathbf{X} = \{ \mathbf{x}_t, ..., \mathbf{x}_{t+T-1} \} \in \mathbb{R}^{T \times J \times 12}$, we first partition it into two sub-sequences based on joint indices: $\mathbf{X}_u$ for the upper body part and $\mathbf{X}_l$ for the lower body part, both contain root data. We then employ two motion encoders, namely $\mathcal{E}^M_u$ and $\mathcal{E}^M_l$, to model the two sub-sequences. This partitioning strategy can significantly reduce the complexity of modeling the entire human pose manifold, as stated in~\cite{hou2023tptn}. The encoders share a similar architecture to the encoder introduced by Zhang et al.~\cite{zhang2023generating}. As a result, the latent vectors $\mathbf{Z}_u \in \mathbb{R}^{\frac{T}{4} \times 128}$ and $\mathbf{Z}_l \in \mathbb{R}^{\frac{T}{4} \times 64}$ for the upper and lower bodies are formulated as follows:
\begin{equation}
\begin{aligned}
    \mathbf{Z}_u = \mathcal{E}^M_u(\mathbf{X}_u), \mathbf{Z}_l = \mathcal{E}^M_l(\mathbf{X}_l)
\end{aligned}
\label{eq:tpmae_enc}
\end{equation}
We treat $\mathbf{Z}_u$ and $\mathbf{Z}_l$ as our motion priors. Note that the input $\mathbf{X}$ has been aligned with the virtual skeleton $\mathbf{\bar{s}}$. Consequently, the encoders are capable of capturing 3D motion kinematic constraints that are independent of skeleton scales, without the interference of skeleton-scale variations. As detailed in Sec.~\ref{sec:tpmae_eva}, these skeleton-scale-independent motion priors contribute to a reduced MPJPE in the reconstructed motion.}

\hsy{\textbf{Decoders.} On the latent manifold, two decoders, $\mathcal{D}_u$ and $\mathcal{D}_l$, are tasked with decoding $\mathbf{Z}_u$ and $\mathbf{Z}_l$, respectively. Additionally, an aggregation layer $\mathcal{A}$, formed by a 1D convolution layer, is exploited to map the decoded features of body parts to the non-root motion $\hat{\mathbf{X}}_{nr}$. Simultaneously, another decoder, $\mathcal{D}_r$, is responsible for directly decoding the concatenation of $\mathbf{Z}_u$ and $\mathbf{Z}_l$ back into the root motion $\hat{\mathbf{X}}_{r}$. This approach of independently reconstructing root and non-root joints has been proven effective by Li et al.~\cite{siyao2022bailando}. In short, the decoding process can be represented as follows:
\begin{equation}
\begin{aligned}
    \hat{\mathbf{X}}_{nr} = \mathcal{A}(\mathcal{D}_u(\mathbf{Z}_u) \oplus \mathcal{D}_l(\mathbf{Z}_l)), \hat{\mathbf{X}}_{r} = \mathcal{D}_r(\mathbf{Z}_u \oplus \mathbf{Z}_l)
\end{aligned}
\label{eq:tpmae_dec_nr}
\end{equation}
where $\oplus$ represents the concatenation operation. Note that $\hat{\mathbf{X}}_{r}$ only contains the root rotations $\mathbf{r}^q$, the 3D root positions $rz^p$ on the Z-axis and the corresponding velocities $rz^v$. During the inference, the global root translations can be derived from $rz^p$, the coordinates of the given 2D keypoints, and the intrinsic parameters of the camera model.}

\hsy{
\textbf{Training Losses.} The loss function $L^M$ used to train TPMAE contains two terms: the motion reconstruction loss $L_{rec}^M$ and the motion smoothness loss $L_{s}^M$, as follows:
\begin{equation}
    \begin{aligned}
        L^M = L_{rec}^M + L_{s}^M
    \end{aligned}
    \label{eq:tpmae_loss}
\end{equation}

\textit{Motion reconstruction loss} is used to constrain the latent manifold to be well-defined. Since different joints matter differently in the motion, we scale the reconstructed and ground-truth motion data by giving relative importance $\bm{\omega}_{r}$ and $\bm{\omega}_{nr}$ to the root joint and other joints like Holden et al.~\cite{holden2017phase} did. Therefore, $L_{rec}^M$ can be computed as:
\begin{equation}
    \begin{aligned}
        L_{rec}^M =  L_1(\bm{\omega}_{r}\hat{\mathbf{X}}_{r},\bm{\omega}_{r}\mathbf{X}_{r}) +  L_1(\bm{\omega}_{nr}\hat{\mathbf{X}}_{nr},\bm{\omega}_{nr}\mathbf{X}_{nr})
    \end{aligned}
    \label{eq:mo_rec}
\end{equation}
where $L_1$ represents the smooth L1 loss function. Specifically, we set the relative importance for the root to be 2.0, the end-effectors to be 1.5, and all other joints to be 1.0.

\textit{Smoothness loss} plays a vital role in preventing the reconstructed motion from abrupt variations along the temporal axis. Furthermore, it bestows added significance to the root, resulting in the following formulation:
\begin{equation}
    \begin{aligned}
        L_{s}^M = & L_1(\hat{\bm{\omega}_{r}\mathbf{Vel}}_{r},\bm{\omega}_{r}\mathbf{Vel}_{r}) + L_1(\hat{\mathbf{Vel}}_{nr},\mathbf{Vel}_{nr})+\\
        & L_1(\bm{\omega}_{r}\hat{\mathbf{Acc}}_{r},\bm{\omega}_{r}\mathbf{Acc}_{r}) + L_1(\hat{\mathbf{Acc}}_{nr},\mathbf{Acc}_{nr})
    \end{aligned}
    \label{eq:tpma_smooth}
\end{equation}
where $\mathbf{Vel}$ \& $\hat{\mathbf{Vel}}$ and $\mathbf{Acc}$ \& $\hat{\mathbf{Acc}}$ represent the velocities and accelerations of the ground-truth \& reconstructed data. Take the $t_{th}$ frame for example, its velocity $\mathbf{vel}_t = \mathbf{x}_t - \mathbf{x}_{t-1}$, and acceleration $\mathbf{acc}_t = \mathbf{vel}_t - \mathbf{vel}_{t-1}$. 
}

\subsection{Learn Human Motion from Videos}
\label{sec:v2m}
\hsy{
After learning the motion priors, we introduce a two-part visual encoder (TPVE) to map the visual features into the learned motion priors. As illustrated in Fig.~\ref{fig:overview}, the TPVE consists of a video feature extractor, two 2D keypoints feature extractors for the upper and lower body parts, two visual feature fusion blocks to fuse the video features and keypoints features, two streams of visual encoders to map the fused visual features into the latent manifold, and a bone ratio predictor to predict the bone scale. 
}

\hsy{
\textbf{Video Feature Extractor.} We utilize the pre-trained ResNet18~\cite{7780459} as our video feature extractor, where the last fully connected layer is discarded. The weights of the pre-trained ResNet18 are fixed during the training procedure. We also employ two learnable 1D convolutional layers, $conv_u$, and $conv_l$, to facilitate our TPVE with the ability to adjust the video features.
}

\hsy{
\textbf{2D Keypoints Feature Extractors.} Similar to the process of motion data, we first divide the 2D keypoint sequences $\mathbf{K} \in \mathbb{R}^{T \times J \times 4}$ into two sub-sequences: $\mathbf{K}_u$ for the upper body part and $\mathbf{K}_l$ for the lower body part. Subsequently, we employ two three-layer CNNs, $\mathcal{K}_u$ and $\mathcal{K}_l$, for these two body parts to extract the 2D keypoint features. 
}

\hsy{
\textbf{Visual Feature Fusion Blocks.} After obtaining the video features and keypoints features, we adopt two residual blocks $\mathcal{F}_u$ and $\mathcal{F}_l$ structured with 1D convolution layers to fuse them, producing the visual features $\tilde{\mathbf{V}}_u$ and $\tilde{\mathbf{V}}_l$:
\begin{equation}
    \begin{aligned}
        \tilde{\mathbf{V}}_u &= \mathcal{F}_u(conv_u(\mathbf{V}) \oplus \mathcal{K}_u(\mathbf{K}_u)) \\
        \tilde{\mathbf{V}}_l &= \mathcal{F}_l(conv_l(\mathbf{V}) \oplus \mathcal{K}_l(\mathbf{K}_l))
    \end{aligned}
    \label{eq:vfb}
\end{equation}
}

\hsy{
\textbf{Visual Encoders.} When we obtain the visual features, we utilize two encoders, $\mathcal{E}^V_u$ and $\mathcal{E}^V_l$, for the upper and lower body parts, to map these visual features into the latent manifold. These encoders have structures similar to the motion encoders but followed by two cross-temporal context aggregation modules (CTCA)~\cite{guo2023distilling}, to better capture the temporal correlations. Then, the latent manifold can be encoded by:
\begin{equation}
    \begin{aligned}
        \tilde{\mathbf{Z}}_u = \mathcal{E}^V_u(\tilde{\mathbf{V}}_u), 
        \tilde{\mathbf{Z}}_l = \mathcal{E}^V_l(\tilde{\mathbf{V}}_l)
    \end{aligned}
    \label{eq:ve_enc}
\end{equation}
where $\tilde{\mathbf{Z}}_u \in \mathbb{R}^{\frac{T}{4} \times 128}$ and $\tilde{\mathbf{Z}}_l \in \mathbb{R}^{\frac{T}{4} \times 64}$ represent the latent manifold encoded from the visual features.
}

\begin{figure*}[t]
  \centering
  \includegraphics[width=0.95\linewidth]{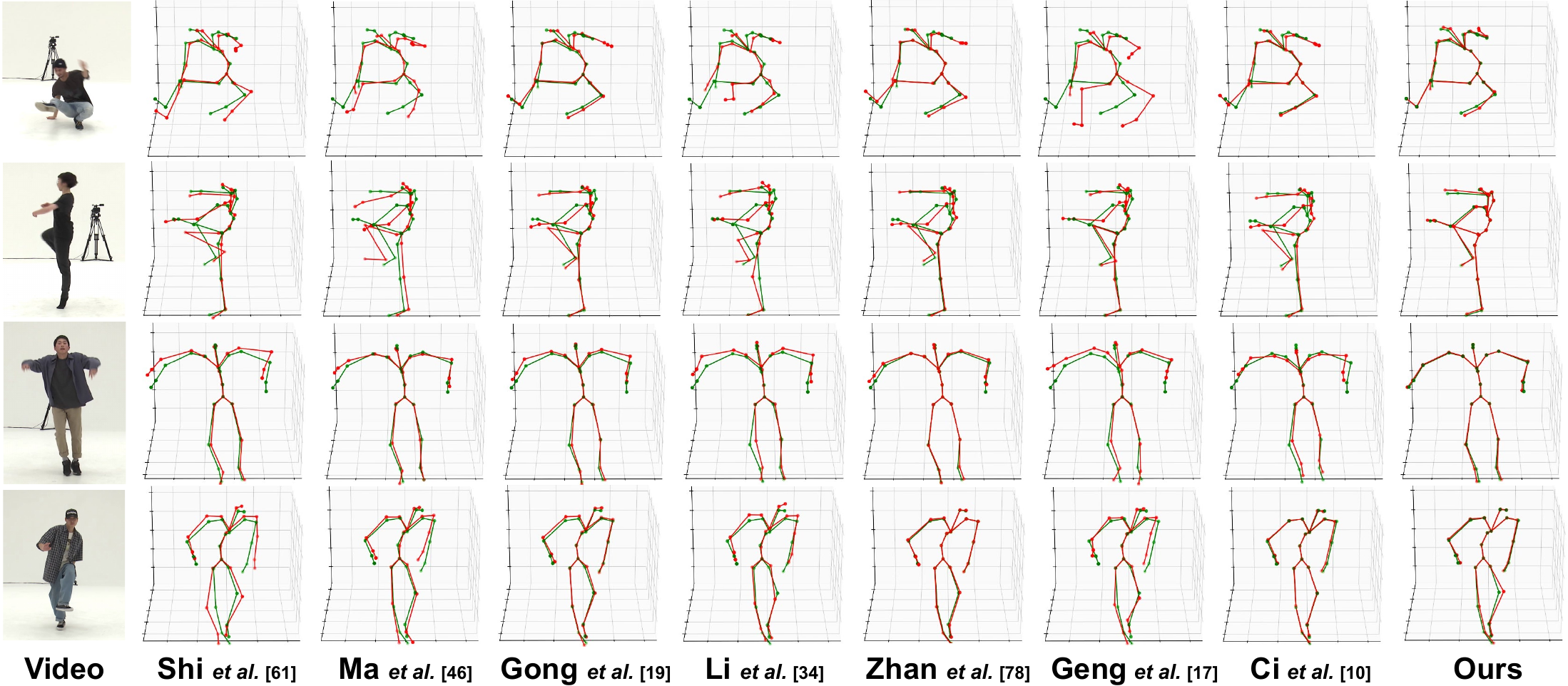}
  \caption{Qualitative comparisons to other SOTA methods. The green skeletons represent the ground truth poses, and the red ones represent the reconstructed poses by different methods.}
  \label{fig:comp1}
\end{figure*}

\hsy{
\textbf{Bone Ratios Predictor.} The scale of the skeleton plays a crucial role in 3D human motion due to substantial variations in human height and proportions for different body parts. While we deliberately remove scale information from the training motion data to enhance the learning of motion priors, we can effectively extract bone ratios $\mathbf{b}$ from the 2D inputs. These inputs inherently embody the scale information of the character. To accomplish this, we utilize another encoder $\mathcal{E}^B$ bearing a similar structure as the motion encoders to predict the bone ratio as follows:
\begin{equation}
    \begin{aligned}
        \tilde{\mathbf{b}} = \mathcal{E}^B(\tilde{\mathbf{V}}_u \oplus \tilde{\mathbf{V}}_l)
    \end{aligned}
    \label{eq:br_enc}
\end{equation}
during inference, the skeleton can be reconstructed by $\tilde{\mathbf{b}} * \bar{s}$.
}
\hsy{
\textbf{Training Losses.} To empower the TPVE with the capability to encode visual features into motion priors, we employ a manifold alignment loss $L_{ma}$ and a bone prediction loss $L_b$ along with the motion prediction and smoothness loss. The TPVE is jointly trained with the TPMAE, resulting in the final loss function of $L$ as follows:
\begin{equation}
    \begin{aligned}
        L = L_{ma} + L_b + L_{pred}^V + L_{s}^V + L^M
    \end{aligned}
    \label{eq:ve_loss}
\end{equation}


\textit{Manifold alignment loss} aims to align the visual latent manifold with the motion priors. To achieve this, we use the smooth L1 loss which is calculated as follows:
\begin{equation}
    \begin{aligned}
        L_{ma} = L_1(\tilde{\mathbf{Z}}_u, \mathbf{Z}_u) + L_1(\tilde{\mathbf{Z}}_l, \mathbf{Z}_l)
    \end{aligned}
    \label{eq:ma_loss}
\end{equation}
Although this method is simple, it surprisingly performs well in aligning the cross-modal latent manifolds for our task. We will further discuss its efficacy in Sec.~\ref{sec:vtm_eva}.

\textit{Bone prediction loss} is a smooth L1 loss as follows:
\begin{equation}
    \begin{aligned}
        L_b = L_1(\hat{\mathbf{b}}, \mathbf{b})
    \end{aligned}
    \label{eq:bone_loss}
\end{equation}

\textit{Motion prediction and smoothness loss} are the same as those employed in TPMAE, which ensures the visual latent vectors can be decoded into the motion data. They can be computed as follows:
\begin{equation}
    \begin{aligned}
        L_{pred}^V = L_1(\bm{\omega}_{r}\tilde{\mathbf{X}}_{r},\bm{\omega}_{r}\mathbf{X}_{r}) + L_1(\bm{\omega}_{nr}\tilde{\mathbf{X}}_{nr},\bm{\omega}_{nr}\mathbf{X}_{nr})
    \end{aligned}
    \label{eq:ve_rec}
\end{equation}
\begin{equation}
    \begin{aligned}
        L_{s}^V = & L_1(\bm{\omega}_{r}\tilde{\mathbf{Vel}}_{r},\bm{\omega}_{r}\mathbf{Vel}_{r}) + L_1(\tilde{\mathbf{Vel}}_{nr},\mathbf{Vel}_{nr})+\\
        & L_1(\bm{\omega}_{r}\tilde{\mathbf{Acc}}_{r},\bm{\omega}_{r}\mathbf{Acc}_{r}) + L_1(\tilde{\mathbf{Acc}}_{nr},\mathbf{Acc}_{nr})
    \end{aligned}
    \label{eq:ve_smooth}
\end{equation}
where $\tilde{\mathbf{X}}_{r}$ and $\tilde{\mathbf{X}}_{nr}$ are the root and non-root data reconstructed from the visual inputs by the decoders. Similarly, $\tilde{\mathbf{Vel}}$ and $\tilde{\mathbf{Acc}}$ represent the velocities and accelerations of the predicted motion.}

\hsy{After training, the motion encoders are discarded, and we only feed the VTM with visual features to generate human motion. The detailed architectures and parameters of all the modules of TPMAE and TPVE will be reported in the supplementary material.}

\section{Experiments}
\label{sec:experiments}

\hsy{\subsection{Implementation Details} 
We implement our algorithm using PyTorch 1.10.1 on a desktop PC equipped with two GeForce RTX 3090 24GB graphics cards and an Intel(R) Xeon(R) E5-2678 CPU with 128GB RAM.

In our experimental setup, the joint number $J$ of the skeleton is set to be 24. The training sequences for the VTM consist of 32 frames each, consecutively extracted from the dataset using a sliding window with a length of 4. For TPMAE, we employ the AdamW optimizer~\cite{loshchilov2017decoupled} for training, running for 500 epochs with a batch size of 100. The initial learning rate is set at 1e-4, subject to decay by multiplication with 0.5 every 100 epochs. Subsequently, the TPVE undergoes joint training with TPMAE using the same settings, except for a reduced batch size of 64.

}

\begin{table}[t]
  \centering
  \resizebox{0.75\linewidth}{!}{
    \begin{tabular}{lcccc}
    \toprule[1pt]
    Method & MPJPE$\downarrow$   & PA-MPJPE$\downarrow$   & MRPE$\downarrow$ & MBLE$\downarrow$ \\ \midrule
    \Shi    & 37.0          & 33.4          & \cellcolor{snd}{45.3} & \cellcolor{snd}{1.4}    \\
    \Ma     & \cellcolor{trd}{20.1} & \cellcolor{snd}{14.8}    & - & 10.4             \\
    \Gong   & 24.8          & 19.6          & -  & 4.8            \\
    \Li     & 45.1          & 37.4          & -  & 9.5            \\
    \Zhan   & \cellcolor{fst}{13.2} & \cellcolor{fst}{10.8} & \cellcolor{trd}{141.4} & \cellcolor{trd}{3.5} \\
    \Geng   & 23.7          & 17.8          & -            & 5.4  \\
    \Ci     & 27.3          & 19.2          & -            & 6.0  \\ \midrule
    Ours   & \cellcolor{snd}{17.8}    & \cellcolor{trd}{15.7} & \cellcolor{fst}{16.8} & \cellcolor{fst}{0.1} \\ \bottomrule[1pt]
    \end{tabular}
  }
  \caption{Comparisons to other SOTA 3D HPE methods on MPJPE, PA-MPJPE, MRPE, and MBLE. The best results are highlighted as \colorbox{fst}{1st}, \colorbox{snd}{2nd}, and \colorbox{trd}{3rd}.}
  \label{tab:comps}
\end{table}

\subsection{Comparisons}
\label{sec:comparison}
\hsy{To the best of our knowledge, MotioNet~\cite{shi2020motionet} is the only existing method that produces a complete 3D human motion representation from 2D input. For a more comprehensive evaluation, we also compare VTM with other SOTA 3D HPE methods that employ different model structures and/or inputs only to predict 3D joint positions. For fair comparisons, we utilize these methods' released codes and training settings to train their models on the AIST++ dataset. Note that the input 2D keypoints for these methods are projected from the corresponding 3D joint positions.} \thy{Please refer to the supplementary material for the detailed settings of these methods.}


\begin{figure*}[t]
  \centering
  \includegraphics[width=0.95\linewidth]{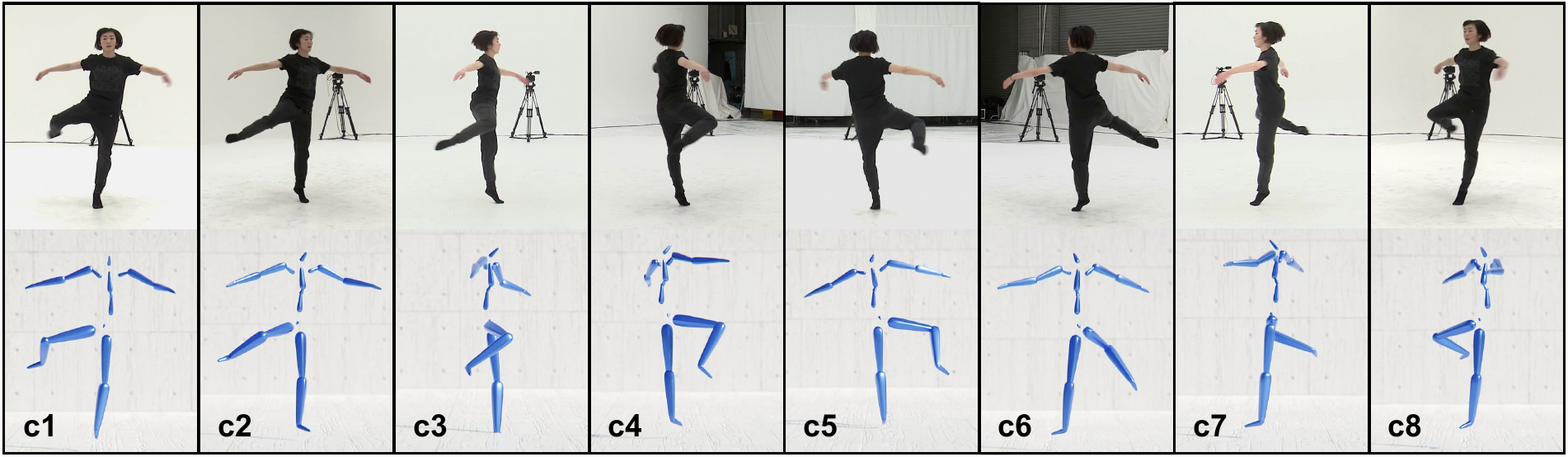}
  \caption{VTM can generalize to different view angles. The first row is the video frames from different camera settings, and the second row is the same pose viewed from the angles corresponding to the camera settings. Only videos and 2D keypoints from camera setting 1, namely c1, are used for training our VTM.}
  \label{fig:multiview}
\end{figure*}

\textbf{Quantitative comparison.} 
\hsy{We employ MPJPE (without rigid alignment) and PA-MPJPE (with rigid alignment), following the common practice in this field, for a fair comparison of joint position reconstruction accuracy. We also employ the mean of 3D bone length error (MBLE) to evaluate the benefits of reconstructing the complete motion representation. Additionally, we assess the mean root position error (MRPE) in comparison to MotioNet and Ray3D~\cite{Zhan_2022_CVPR}, as these two methods reconstruct global root translations as well. All metrics are in millimeters and computed on the validation dataset .}

\hsy{Since MotioNet and VTM generate the joint rotations, we perform FK using the reconstructed skeletons and joint rotations to obtain 3D joint positions for metric computation. In Tab.~\ref{tab:comps}, VTM ranks second and third in terms of MPJPE and PA-MPJPE. This can be attributed to VTM's slightly different objective, focusing on reconstructing a complete 3D motion representation from monocular videos, setting it apart from methods primarily concentrating on 3D joint positions. Notably, VTM excels in MRPE compared to MotioNet and Ray3D, benefiting from encoding global root positions into motion priors along with the joint rotations. Additionally, VTM and MotioNet secure the top two positions in MBLE, showcasing the effectiveness of reconstructing complete motion representation in mitigating bone length inconsistency issues encountered when reconstructing 3D joint positions.} 

\hsy{We train VTM on the Human3.6M dataset and achieve the three metrics on the test dataset (S9 and S11) as (67.3, 49.1, 15.6). It's worth noting that the Human3.6M skeleton includes three additional joints corresponding to zero-length bones, which are crucial for skeleton-and-rotation-driven human motion. However, these joints may bring complexity during network optimization. If our focus is solely on 3D joint positions, the issue of rotation confusion becomes less relevant. This contextualizes why VTM's MPJPE metrics might not match those of SOTA methods like Ray3D, but the MRPE surpasses Ray3D's reported value of 109.5mm.}

\begin{table}[t]
  \centering
  \resizebox{0.5\linewidth}{!}{
    \begin{tabular}{lccc}
    \toprule[1pt]
    Method       & MPJPE$\downarrow$  & PA-MPJPE$\downarrow$  & MRPE$\downarrow$ \\ \midrule
    TPMAE-SKW & 4.9          & 4.1          & 0.8           \\
    TPMAE-Q   & 84.0         & 10.7         & 98.5          \\
    OPMAE        & 5.7          & 4.9          & 1.1           \\ \midrule
    TPMAE        & {4.8} & {4.0} & {0.8}  \\ \bottomrule[1pt]
    \end{tabular}
  }
  \caption{Evaluations of TPMAE on MPJPE, PA-MPJPE, and MRPE.}
  \label{tab:tpmae_abla}
\end{table}


\hsy{
\textbf{Qualitative comparison.} To visually compare the quality of reconstructed motion among different methods, we randomly select four video frames from the validation set and show the reconstructed 3D joint positions in Fig.~\ref{fig:comp1}. Notably, VTM consistently surpasses or is on par with other methods. Despite quantitative and qualitative weaknesses compared to Ray3D in many cases, VTM exhibits superiority when significant self-occlusions occur in the video, exemplified in the second row of Fig.~\ref{fig:comp1}. We attribute this advantage to our motion priors, which encode reasonable 3D motion manifolds. This effective encoding captures correlations between joint rotations, enabling the inference of plausible human body poses in such situations.}

\subsection{Evaluations}
\label{sec:ablation}
\hsy{
\subsubsection{Evaluations of TPMAE}
\label{sec:tpmae_eva}
We conduct experiments to validate our TPMAE's efficacy through various configurations: 1) TPMAE-SKW: In this scenario, TPMAE is trained using motion data in original skeletons instead of those aligned with $\bar{s}$. This investigation aims to discern the impact of skeleton scales on motion reconstruction accuracy. 2) TPMAE-Q: Using quaternion-based rotation and a local 3D joint position representation similar to that used in~\cite{holden2017phase, hou2023tptn}. 3) OPMAE: To assess the necessity of the two-part design in TPMAE, we discard the two-part structure and increase the latent space dimension to 196. This adjustment is made to keep the size of the latent manifold unchanged.

We perform forward kinematics using the predicted joint rotations and ground-truth skeletons, presenting the quantitative results in Tab.~\ref{tab:tpmae_abla}. The superiority of TPMAE over its variants on these metrics confirms its effectiveness. In TPMAE-Q, representing root rotations as angular velocities around the Y-axis leads to the accumulation of errors, especially in long sequences. This explains why TPMAE-Q performs comparably on PA-MPJPE but poorly on MPJPE compared to other models.
}

\begin{figure}[t]
  \centering
  \includegraphics[width=0.95\linewidth]{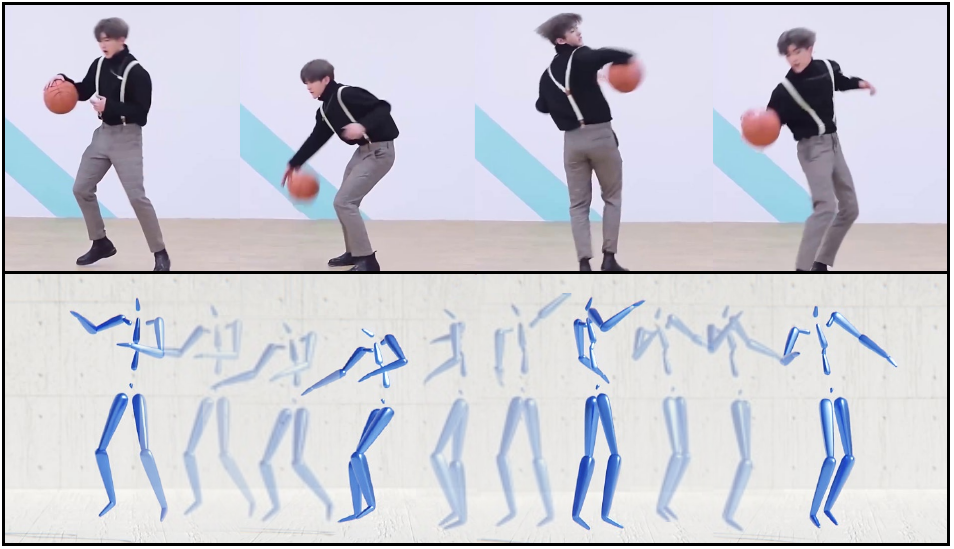}
  \caption{VTM can reconstruct motion from in-the-wild videos. The first row is the continuous frames from a wild video, and the second row shows the corresponding results produced by VTM.}
  \label{fig:wild_video}
\end{figure}

\subsubsection{Evaluations of VTM}
\label{sec:vtm_eva}

\hsy{
\textbf{Generalization to unseen view angles.} As Sec.~\ref{sec:data_preparation} mentions, our VTM is exclusively trained using the 2D keypoints and videos obtained from camera 1. In this particular experiment, we evaluate the performance of our VTM by utilizing the 2D data collected from cameras 2 to 8 in the AIST++ dataset. As illustrated in Fig.~\ref{fig:multiview}, our VTM can generalize well across previously unseen view angles. 

\textbf{Generalization to in-the-wild videos.} We introduce a mapper network to enable VTM to reconstruct motion from in-the-wild videos. This network, constructed as a four-layer MLP, is designed to convert COCO-formatted keypoints to our virtual 2D keypoints. This mapper facilitates the seamless application of our VTM to in-the-wild videos, so that we can directly utilize 2D keypoints detected by readily available 2D HPE models. As illustrated in Fig.~\ref{fig:wild_video} and e) \& f) of Fig.~\ref{fig:teaser}, the VTM consistently produces plausible 3D human motion from in-the-wild videos.
}

\begin{table}[t]
  \centering
  \resizebox{0.6\linewidth}{!}{
    \begin{tabular}{lccc}
    \toprule[1pt]
    Method            & MPJPE$\downarrow$   & PA-MPJPE$\downarrow$   & MRPE$\downarrow$          \\ \midrule
    VTM-CL            & 22.5          & 19.4          & 21.1          \\
    VTM-CL+L1         & 20.5          & 17.8          & \cellcolor{snd}{19.2}    \\
    VTM-SKW        & \cellcolor{snd}{19.1}    & \cellcolor{snd}{16.8}    & 22.2          \\
    VTM-OP            & 72.1          & 58.8          & 65.2          \\
    VTM-w/o\_JT       & 21.8          & 18.1          & 24.3          \\
    VTM-w/o\_PT & 20.9          & 18.5          & 21.3          \\
    VTM-w/o\_CTCA     & 21.0          & 18.4          & \cellcolor{trd}{20.8} \\
    VTM-w/o\_K      & 73.5          & 59.7          & 66.5          \\
    VTM-w/o\_V      & \cellcolor{trd}{20.2} & \cellcolor{trd}{17.5} & 31.8          \\ \midrule
    VTM              & \cellcolor{fst}{17.8} & \cellcolor{fst}{15.7} & \cellcolor{fst}{16.8} \\ \bottomrule[1pt]
    \end{tabular}
  }
  \caption{Evaluations of VTM on MPJPE, PA-MPJPE, and MRPE. The best results are highlighted as \colorbox{fst}{1st}, \colorbox{snd}{2nd}, and \colorbox{trd}{3rd}.}
  \label{tab:vtm_abla}
\end{table}

\hsy{
\textbf{Evaluations on motion priors.} Our VTM leverages pre-learned motion priors by aligning the visual latent manifold with them, undergoing joint training with the pre-trained TPMAE. To assess its effectiveness, we conducted the following experiments.

\textit{VTM-CL and VTM-CL+L1.} These experiments aim to identify a suitable method for aligning cross-modal latent manifolds. In VTM-CL, we replaced $L_{ma}$ with a contrastive loss (CL), similar to the methodology used in CLIP. This loss function is calculated based on scaled pairwise cosine similarities between visual latent vectors and motion prior vectors. In VTM-CL+L1, we integrated the CL loss into $L_{ma}$. However, as indicated in Tab.~\ref{tab:vtm_abla}, both variants are inferior to our VTM. We posit that the CL loss may be unsuitable for our task due to the presence of numerous similar poses in the dataset. Selecting training batches in a sliding window manner can introduce sequences with higher similarity. Simply treating such similar sequences in a batch as negative samples may confuse the optimization of the network.

\textit{VTM-w/o\_JT and VTM-w/o\_PT.} To explore joint training with the pre-trained TPMAE, we conduct two additional experiments: 1) training only the TPVE while keeping the weights of TPMAE decoders fixed (VTM-w/o\_JT), and 2) directly training TPMAE and TPVE end-to-end without pre-training TPMAE (VTM-w/o\_PT). As demonstrated in Tab.~\ref{tab:vtm_abla}, the joint training strategy can significantly improve the accuracy of motion and global root translation reconstruction.

\textbf{Other Evaluations.} The efficacy of various configurations of our VTM is assessed through a series of experiments, with quantitative results reported in Tab.~\ref{tab:vtm_abla}.

\textit{VTM-SKW.} Utilizing TPMAE-SKW as the motion auto-encoder, this model is trained on motion data in original skeletons to examine the impact of skeleton scales on 3D motion reconstruction. The results demonstrate that aligning human motion to our virtual skeleton significantly enhances motion reconstruction from monocular videos. By excluding skeleton scale information from the training data, VTM focuses solely on learning kinematic constraints such as joint rotations, which proves advantageous for our task.

\textit{VTM-OP.} We abandon the two-part strategy and use the OPMAE as the motion auto-encoder to analyze the advantages of our two-part design. Despite the comparable performance of OPMAE to TPMAE, the VTM-OP lags behind VTM by a large margin, indicating that learning human motion from monocular videos is far more challenging than motion reconstruction.

\textit{VTM-w/o\_CTCA.} This model is trained with the exclusion of the CTCA modules from the visual encoders. Tab.~\ref{tab:vtm_abla} demonstrates that CTCA plays a crucial role in enhancing the temporal correlations captured by VTM, resulting in higher motion reconstruction accuracy.

\textit{VTM-w/o\_K and VTM-w/o\_V.} We investigate how different 2D inputs influence VTM performance. When fed only with videos (VTM-w/o\_K), there is a significant decrease in performance across all metrics. Conversely, when provided with only 2D keypoints (VTM-w/o\_V), VTM-w/o\_V maintains top-3 performance in MPJPE but experiences a notable drop in MRPE. This suggests that 2D keypoints assist the model in identifying motion contours, while video information contributes to global spatial awareness.

}
\section{Conclusion}
\label{sec:conclusion}

\hsy{
In conclusion, we present an innovative framework named VTM, specializing in learning human motion from monocular videos. Our VTM first learns motion priors from 3D motion data and subsequently leverages these well-defined priors to reconstruct high-fidelity motion from 2D inputs. We conduct comprehensive experiments to validate the effectiveness of our proposed method. Notably, VTM exhibits satisfactory generalization capabilities across various unseen view angles and performs robustly on in-the-wild videos.
}

\hsy{
\textbf{Future work.} Most high-quality human motion datasets acquired by professional motion capture systems, such as CMU mocap~\cite{cmu_mocap}, are unfortunately devoid of paired video data, rendering them unsuitable for our current supervised learning approach. In the future, we plan to investigate unsupervised or semi-supervised learning methods. This exploration seeks to leverage existing high-quality human motion datasets alongside readily available video data, intending to enhance the model's generalization capabilities and robustness.
}

{
    \small
    \bibliographystyle{ieeenat_fullname}
    \bibliography{main}
}



\end{document}